\documentclass{article}

\usepackage[preprint]{neurips_2025}

\usepackage[utf8]{inputenc} 
\usepackage[T1]{fontenc}    
\usepackage{hyperref}       
\usepackage{url}            
\usepackage{booktabs}       
\usepackage{amsfonts}       
\usepackage{xspace}         
\usepackage{nicefrac}       
\usepackage{microtype}      
\usepackage{xcolor}         

\newcommand{\eg}{\emph{e.g.}\xspace}

\newcommand{\eat}[1]{}

\newcommand{\better}[1]{\textcolor{magenta}{#1}}
\usepackage{microtype}
\usepackage{graphicx}
\usepackage{subfigure}
\usepackage{booktabs} 
\usepackage{hyperref}
\usepackage{amsmath}
\usepackage{amssymb}
\usepackage{mathtools}
\usepackage{amsthm}
\usepackage{multirow}
\usepackage{array}
\usepackage{wrapfig}
\usepackage{caption}

\usepackage{algorithm}
\usepackage{algorithmic}
\usepackage{amsmath}
\usepackage{amssymb}
\usepackage{enumitem}

\title{PhyDA: Physics-Guided Diffusion Models for Data Assimilation in Atmospheric Systems}

\author{
  Hao Wang$^{1}$, Jindong Han$^{3}$, Wei Fan$^{4}$, Weijia Zhang$^{1}$, Hao Liu$^{1,2}$ \\
  $^{1}$Hong Kong University of Science and Technology (Guangzhou)\\
  $^2$Hong Kong University of Science and Technology \\
  $^3$Shandong University\\
  $^4$University of Oxford\\
\texttt{\{figerhaowang, hanjindong01, weifan.oxford, vegazhang3\}@gmail.com,}
\\ \texttt{liuh@ust.hk} 
}

\begin{document}

\maketitle

\begin{abstract}

Data Assimilation (DA) plays a critical role in atmospheric science by reconstructing spatially continous estimates of the system state, which serves as initial conditions for scientific analysis.
While recent advances in diffusion models have shown great potential for DA tasks, most existing approaches remain purely data-driven and often overlook the physical laws that govern complex atmospheric dynamics. As a result, they may yield physically inconsistent reconstructions that impair downstream applications.
To overcome this limitation, we propose \textbf{PhyDA}, a physics-guided diffusion framework designed to ensure physical coherence in atmospheric data assimilation. PhyDA introduces two key components: (1) a \textit{Physically Regularized Diffusion Objective} that integrates physical constraints into the training process by penalizing deviations from known physical laws expressed as partial differential equations, and (2) a \textit{Virtual Reconstruction Encoder} that bridges observational sparsity for structured latent representations, further enhancing the model's ability to infer complete and physically coherent states. Experiments on the ERA5 reanalysis dataset demonstrate that PhyDA achieves superior accuracy and better physical plausibility compared to state-of-the-art baselines. Our results emphasize the importance of combining generative modeling with domain-specific physical knowledge and show that PhyDA offers a promising direction for improving real-world data assimilation systems. 
\end{abstract}

\section{Introduction}

Understanding and forecasting complex dynamical systems lies at the heart of many scientific domains, especially atmospheric and oceanic sciences~\citep{lorenc1986analysis, evensen1994sequential, carrassi2018data}. These systems evolve over space and time under the influence of nonlinear physical laws, making it essential to estimate their internal state as accurately as possible. Data Assimilation (DA) addresses this challenge by combining numerical model predictions (referred to as the background) with sparse and noisy real-world observations to produce a more accurate estimate of the spatially continous system state, known as the analysis. This process improves both short-term forecasts and scientific understanding, and is widely used in weather prediction and climate reanalysis workflows.

Traditional DA techniques, such as variational methods (\eg 3D-Var, 4D-Var) \citep{le1986variational, courtier1998ecmwf, tr2006accounting} and ensemble-based approaches (\eg EnKF) \citep{bannister2017review}, have formed the backbone of operational forecasting for decades. However, these methods often assume linear dynamics, Gaussian error distributions, and fixed covariance structures, limiting the ability to capture the nonlinear and multiscale nature of atmospheric processes. Additionally, they often require adjoint models or large ensembles~\citep{wang2021physics}, which is computationally prohibitive in practice. To overcome these limitations, recent work has explored deep generative models as alternatives for DA~\citep{chen2023fengwu, lam2023learning}, with a particular focus on diffusion models \citep{ho2020denoising, dhariwal2021diffusion, gupta2024photorealistic}. These models excel at modeling complex, high-dimensional distributions and are able to efficiently reconstruct atmospheric states from sparse observations \citep{huang2024diffda, rozet2023score}.

Despite recent progress, existing diffusion-based DA methods generally lack built-in mechanisms to enforce physical consistency between tightly coupled atmospheric variables such as temperature, pressure, and humidity. In real-world systems, these variables are governed by physical laws including conservation laws and thermodynamic principles~\citep{holton2012dynamic, wallace2006atmospheric}. Without explicit constraints, purely data-driven models may generate outputs that violate physical relationships, resulting in unrealistic or unstable states~\citep{brandstetter2022clifford, karniadakis2021physics, rasp2020weatherbench}. Moreover, the inherent sparse nature of real-world atmospheric observations further complicate this problem. Many physical constraints require access to gradients or multivariable dependencies across space, which are difficult to compute or infer from limited and unevenly distributed observations.

\begin{wrapfigure}{r}{0.5\textwidth}  
  \centering
  \vspace{-10pt}  
  \includegraphics[width=0.5\textwidth]{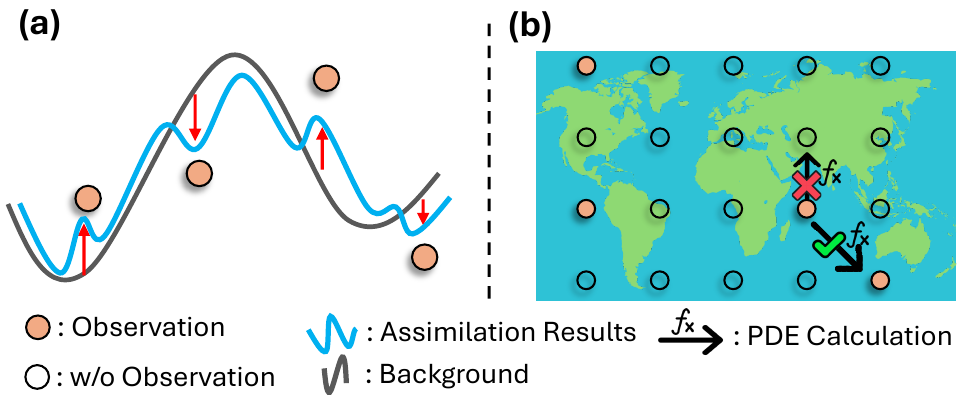}
  \caption{Figure (a) provides a systematic framework that calibrates background states with sparse observations to produce analysis. Figure (b) shows the challenge caused by observational sparsity. The absence of neighbor observations hinders the calculation of spatial gradients which is common in physical PDE calculation.}
  \label{fig:intro}
  \vspace{-5pt}  
\end{wrapfigure}

To tackle the above challenges, we propose PhyDA, a physics-guided diffusion framework for atmospheric DA. PhyDA is composed of two complementary modules: a Physically Regularized Diffusion Objective (PRDO) and a Virtual Reconstruction Encoder (VRE). PRDO embeds physical constraints, formulated as partial differential equations (PDEs), into the training loss, augmenting the standard score-matching objective with physics-based penalties. This ensures that the generative process respects the governing dynamics of atmospheric systems, rather than relying on post-hoc correction \citep{ruhling2024probablistic}. While PRDO provides a strong physical prior, it alone is insufficient in the face of highly sparse observations. To this end, we further introduce VRE, which transforms scattered and noisy observations into a structured latent space that captures multiscale structures and cross-variable dependencies. This latent representation serves as a coarse-grained structured prior to condition the diffusion process, enabling physically coherent assimilation even in under-observed regions.

Together, PRDO and VRE allow PhyDA to seamlessly integrate data-driven generative modeling with domain-specific knowledge, leading to reconstructions that are not only statistically accurate but also physically faithful. Experiments on the ERA5 reanalysis dataset show that PhyDA consistently outperforms state-of-the-art methods in both numerical accuracy and physical realism, demonstrating its potential for deployment in real-world atmospheric modeling systems.
Our main contributions are summarized as follows:
\begin{itemize}[leftmargin=*]
    \item We formulate data assimilation as a physics-informed generative modeling task, emphasizing the preservation of physical relationships among atmospheric variables.
    \item We introduce PhyDA, a physics-guided framework that embeds PDE-based atomspheric constraints into the diffusion process through a principled training objective, and leverages a latent encoder to encode sparse observations into coarse-grained spatial representations as condition signals.
    \item We evaluate PhyDA on the ERA5 reanalysis dataset and show that it outperforms existing baselines in both reconstruction accuracy and physical consistency.
\end{itemize}

\begin{figure*}[t]
  \centering
  \includegraphics[width=\linewidth]{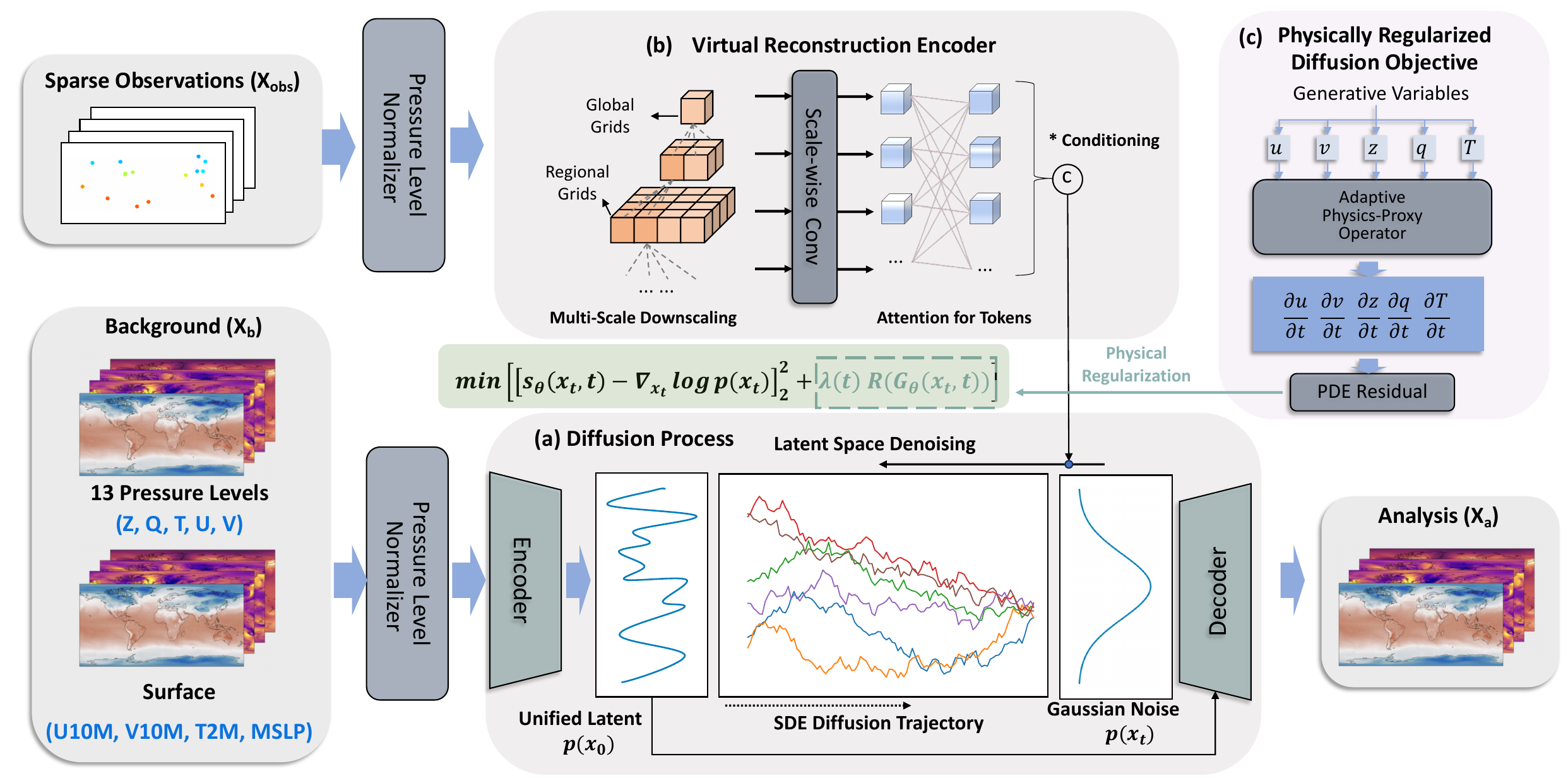}
  \centering
  \caption{Overview of the PhyDA framework. Sparse observations and gridded atmospheric variables (background) are first normalized. (a) The background is further encoded by a variational autoencoder into a unified latent space where a score-based diffusion model simulates stochastic trajectories. (c) During training, a Physically Regularized Diffusion Objective enforces consistency with physical laws by minimizing PDE residuals while an Adaptive Physics-Proxy Operator is adopted to simplify the calculation of partial derivative in complex PDEs. (b) A Virtual Reconstruction Encoder extracts hierarchical features from sparse inputs for conditioning the diffusion process.  The model reconstructs physically consistent analysis states that align with both observations and governing physical PDEs.}
  \label{fig:framework}
  \vspace{-15pt}
\end{figure*}

\section{Preliminary}

\subsection{Problem Formulation}
Data assimilation (DA) aims to estimate the full state of an atmospheric system \( x_i \in \mathbb{R}^n \) at a fixed time step \( i \), given a limited number of sparse observations \( y_i \in \mathbb{R}^m \) and a prior forecast \( \hat{x}_i \in \mathbb{R}^n \) produced by a predictive model \( \mathrm{F}(x_{i-1}) \). 
Here, $m, n$ denote the number of observations and full state, respectively. 
These states \( x_i \) are defined over a grid of spatiotemporal points and include key atmospheric variables such as temperature, wind, pressure, and humidity across multiple pressure levels.
From a probabilistic perspective, the goal of DA is to recover the posterior distribution \( p(x_i \mid \hat{x}_i, y_i) \), which combines prior knowledge from the model with new, typically sparse, observational data. This task is fundamentally ill-posed due to the high dimensionality of \( x_i \), the sparsity \( y_i \), and the need to preserve physical coherence among the variables — a property often ignored by conventional deep learning-based approaches.

In this work, we reformulate data assimilation as a \textit{conditional generative modeling problem}, where we seek to sample physically coherent states from \( p(x_i \mid \hat{x}_i, y_i) \) using a score-based diffusion model. To facilitate this, we encode the inputs as two high-dimensional tensors:  
a \textbf{background field} \( \mathbf{X}_{b} \in \mathbb{R}^{H \times W \times (P \cdot K  + s)} \), which contains prior states over spatial grids, pressure levels, and physical variables, and an \textbf{observation field} \( \mathbf{X}_{obs} \in \mathbb{R}^{H \times W \times (P \cdot K  + s)} \), which includes sparse station-based measurements.  
Here, $H$, $W$ represent the height and width of the spatial grids, $P$ denotes the number of pressure levels, $K$ denotes the number of variables in each pressure, and $s$ denotes the number of surface variables.
Our method leverages both fields to guide the generative process toward accurate and \textit{physically grounded analysis}  \( \mathbf{X}_{a} \in \mathbb{R}^{H \times W \times (P \cdot K  + s)} \). The key technical challenge lies in aligning the generated samples with known physics (e.g., via PDEs) while remaining consistent with limited observations.

\subsection{Score-based Diffusion Models}
\label{sec:score-based}

While Denoising Diffusion Probabilistic Models (DDPMs) \citep{ho2020denoising} discretize the diffusion process over fixed time steps, score-based diffusion models offer a powerful framework for generative modeling by simulating a continuous-time stochastic process, enabling more flexible and theoretically grounded sampling procedures. 
This process samples from noise by first perturbing the data with gradually increasing noise, then learning to reverse the perturbation via estimating the score function, which is the gradient of the log-density function with respect to the data. 
The forward process is defined by a stochastic differential equation (SDE): 
\begin{align}
    dx = f(x(t))dt + g(t)dw(t),
    \label{eq:sde}
\end{align}
where \( f(x(t)) \) and \( g(t) \) are the drift and diffusion coefficient, respectively, and \( w(t) \) represents a Wiener process.
This diffusion process perturbs samples from the data distribution $p(x)$ into noise over time.
To recover data, the model simulates a reverse-time SDE: 
\begin{align}
    dx = [f(x(t)) - g(t)^2 \nabla_{x} \log p_t(x)]dt + g(t)dw(t),
\label{eq:reverse-sde}
\end{align}
which requires estimating the score function $\nabla_{x} \log p(x(t))$.
Since this gradient is intractable, a neural network $s_\theta(x(t), t)$ is trained to approximate it via denoising score matching:
\begin{equation} \label{eq:dsm-objective}
    \arg\min_\phi \mathbb{E}_{p(x) p(t) p(x(t) \mid x)} \left[ \sigma(t)^2 \left\| s_\phi(x(t), t) - \nabla{x(t)} \log p(x(t) \mid x) \right\|_2^2 \right]
\end{equation}
Once trained, the score network guides the reverse SDE to generate high-fidelity samples from the target distribution. For a detailed formulation and training procedure, please refer to Appendix A.

\section{Methodology}

The overall framework and model details of PhyDA are depicted in Figure ~\ref{fig:framework}.
Initially, the background and observations undergo a pressure level normalizer before they are put into the networks. The normalized background is further encoded into a unified latent space by a variational autoencoder. During training process, PhyDA leverages a Physically Regularized Diffusion Objective (PRDO) to learn high-level correlations among multiple physical variables of physical states under the guidance of PDE-governed denoise process. Finally, the denoised latent representation is mapped to the analysis by the decoder.
Moreover, we further condition the model via a Virtual Reconstruction Encoder (VRE), which is specifially designed for capturing the intricate physical relationships of the sparse observations.
In the following subsections, we provide a detailed description of the process for enforcing the physical consistency.

\subsection{Enforcing Physical Coherence within Score-Based Diffusion}

While highly effective for data-driven generation, standard score-based models treat the data manifold as an implicit function learned from samples and ignore known physical principles. In atmospheric modeling, however, data are governed by complex physical laws expressed as partial differential equations (PDEs). Ignoring these constraints can result in physically implausible outputs, especially when observations are sparse or ambiguous.

Incorporating physical knowledge into diffusion models is therefore essential but non-trivial. Simply hard-coding constraints would break the differentiability and flexibility of the learned score function. Moreover, enforcing physical consistency during sampling requires aligning generated trajectories with complex, nonlinear dynamics — a task ill-suited to traditional loss functions.
To address this problem, we take inspiration from the energy-based modeling perspective: the learned score can be interpreted as the gradient of a log-probability combining data likelihood with a physically-informed energy landscape.

\textbf{Theorem 1: Energy-Based Interpretation of Physical Constraints.}
\textit{Let $R(x_0)$ denote the residual derived from the governing equations applied to a generated sample $x_0$, such that $R(x_0) = \mathcal{F}(D^k x_0, \ldots)$ (see Definition 1 in Appendix C).
Define a virtual observation model for the residual as:
\begin{equation}
    q_R(\hat{r} \mid x_0) = \mathcal{N}(\hat{r}; R(x_0), \sigma^2 I),
\end{equation}
with target \( \hat{r} = 0 \). Then, the log-likelihood of satisfying the physical constraints is given by:
\begin{equation}
    \log q_R(0 \mid x_0) = -\frac{1}{2\sigma^2} \| R(x_0) \|^2 + \text{const}.
\end{equation}
In the context of energy-based modeling, the learned score function $\nabla_{x_t}\log p(x_t)$ can be interpreted as the gradient of a log-probability that combines data likelihood with a physically-informed energy landscape. Specifically:
\begin{equation}
    \nabla_{x_t} \log p(x_t) \propto \nabla_{x_t} \log q(x_t) - \lambda \cdot \nabla_{x_t} \mathbb{E}_{x_0 | x_t} \left[ \| R(x_0) \|^2 \right].
\end{equation}}

Thus, minimizing the residual loss is equivalent to maximizing the likelihood of satisfying physical laws. By integrating this term into the generative training objective, we obtain a posterior-weighted sample distribution that favors physically admissible states:
\begin{equation}
    \min \mathbb{E}_{x_t, t} [\underbrace{\left\| s_\theta(x_t,t) - \nabla_{x_t}\log p(x_t) \right\|^2}_{\text{Score Matching}} + \lambda(t) \underbrace{\| R(G_\theta(x_t, t)) \|^2}_{\text{Physics Regularization}} ],
\end{equation}
where $G_\theta$ represents the denoising network that predicts the clean physical state from noisy sample $x_t$.
In this formulation, the physical constraint acts as a soft prior, guiding the generative process toward solutions that not only match the training data but also respect fundamental physical laws. As we will show, this enables more stable, accurate, and physically consistent data assimilation, particularly under sparse observational regimes.

Furthermore, calculating the governing PDEs usually requires the use of differential and integral operations, which are usually computational cost-expensive. 
In order to efficiently calculate such PDEs and enbale loss backward, we propose Adaptive Physics-Proxy Operator, a carefully-designed implemention of differentiation and integration through spectral filtering and selective enforcement, respectively. 
Given the governing equations $\mathcal{F}(D^ku, ..., u, x) = 0$, the partial derivative of each atmospheric variable with respect to time can be separated mathematically \citep{xu2024generalizing}, denoted as $\mathcal{K}_{PDE}$, which takes current weather state $\mathbf{X}_t$ as input and produces the derivative of each variable with respect to time as:
{
\begin{equation}
    \mathcal{K}_{PDE}(\mathbf{X}_t) = \frac{\partial \gamma}{\partial t}= S_\gamma(u,v,z,q,T), \quad \gamma\in \{u,v,z,q,T\}
\end{equation}}

By this way, the complex calculation of PDEs can be simplied thus reduce the computational cost.
This allows for a scalable and computationally feasible approach to integrating physical laws into our generative model, enabling the model to produce physically consistent samples without excessive computational overhead.

\subsection{Bridging Sparsity for Physically Coherent States}

As discussed earlier, the extreme sparsity of observational data poses a significant challenge in modeling intricate physical dependencies, especially among spatially isolated points. Conventional encoders, such as attention mechanisms or convolutional networks, often struggle to learn reliable correlations from such sparse and discontinuous data, leading to difficulties in conditioning the diffusion process.

To address this, we propose Virtual Reconstruction Encoding (VRE), a specialized encoder designed to transform sparse, spatially discontinuous observations into rich, continuous representations. VRE effectively simulates the physical evolution of the system, providing high-quality conditional information to guide the generative process.

Given the input observation \( \mathbf{X}_{obs} \in \mathbb{R}^{H \times W \times (P \times K + s)} \) from a predictive model, we first normalize the input \( \mathbf{X}_{obs} \) ($\mathbf{X}$ for convenience) for each node to have zero mean and unit variance using Reversible Instance Normalization (RevIN) independently across pressure levels:
\begin{equation}
\begin{aligned}
\mu_k = \frac{1}{H  WP} \sum_{i=1}^{H} \sum_{j=1}^W \sum_{p=1}^P \textbf{X}_{i,j,k,p},
\quad \sigma_k^2 = \frac{1}{HWP}\sum_{i=1}^{H} \sum_{j=1}^{W}\sum_{p=1}^P(\mathbf{X}_{i,j}-\mathbb{E}[\mathbf{X}_k])^{2},
\end{aligned}
\end{equation}

where \( \mu_k \) and \( \sigma_k^2 \) are the mean and variance of the atmospheric variable \( k \), respectively. By normalizing the observation \( \mathbf{X}_{obs} \) in a variable-independent manner, we prevent the model from overfitting to the specific characteristics of any individual variable:
{
\begin{equation}
    \mathbf{X}_{\text{norm}} = \operatorname{Concat}\left(\frac{\mathbf{X}_k - \mu_k}{\sigma_k + \epsilon}\right), \quad k \in K,
\end{equation}}

where \( \epsilon \) is a small constant to ensure numerical stability, and \( \operatorname{Concat}(\cdot) \) denotes concatenation over the \( K \) variables.

Next, inspired by the work of \citep{wu2024transolver}, we generate a series of multi-scale virtual tokens by applying a set of sliding windows with varying resolutions. These virtual tokens address the challenge of spatial discontinuity by mapping sparse observation points into a continuous spatial representation. This increases the density of information and enhances the model’s ability to extract meaningful correlations from sparse observations.

The input observational state \( \mathbf{X}_{\text{norm}} \) is progressively downsampled across \( M \) scales using convolution operations with a stride of 2, producing the multi-scale set \( \chi_{\text{obs}} = \{ \mathbf{x}_0, \dots, \mathbf{x}_M \} \), where \( \mathbf{x}_\tau \in \mathbb{R}^{\frac{H \times W}{4^\tau} \times (P \times K + s)} \). The downsampling process follows the recursive relationship:
{
\begin{equation}
    \mathbf{z}_\tau = \operatorname{Conv}(\mathbf{x}_{\tau-1}, \text{stride}=2), \quad \tau \in \{1, \dots, M\}.
\end{equation}}

By applying 2D convolutions to these multi-scale tokens derived from sparse observations, we capture long-range spatial correlations between observation stations, thus enriching the information available for further encoding.

To further extract meaningful information from these multi-scale tokens, we apply an attention mechanism to model intricate correlations across different observational states. Specifically, we compute the attention between encoded tokens as:
$
\mathbf{q}, \mathbf{k}, \mathbf{v} = \operatorname{Linear}(\mathbf{z}),
\quad \mathbf{z}' = \operatorname{Softmax}\left(\frac{\mathbf{q}\mathbf{k}^T}{\sqrt{C}}\right) \mathbf{v},
$
where \( \mathbf{z}' \) represents the updated, attention-weighted encoding of the observational states. This process ensures that the sparse observations are effectively utilized, providing a richer and more accurate conditioning for the diffusion models.
By using VRE, we resolve the problem of spatial discontinuity in observations, enabling the model to leverage sparse data effectively while maintaining high-quality conditional information for more accurate and physically consistent data assimilation.

\section{Experiment}
\subsection{Implementation}
\textbf{Datasets.} 
We use the ECMWF Reanalysis v5 (ERA5) \citep{hersbach2019era5} dataset as the resource part of the training data, a global atmospheric reanalysis archive containing hourly weather variables such as geopotential, 
temperature, wind speed, humidity, etc.
We demonstrate our method in a real-world scenario containing 5 pressure-level variables (temperature, geopotential, u-wind, v-wind,  specific humidity) with a horizontal resolution of 1.5 degree and and 13 vertical levels (50hPa, 100hPa, 150hPa, 200hPa, 250hPa, 300hPa, 400hPa, 500hPa, 600hPa, 700hPa, 850hPa, 925hPahPa, 1000hPa). We use the predictive results from Fengwu model \citep{chen2023fengwu} as the background of our data assimilation tasks.
The dataset contains values for our target atmospheric variables from 1979 to 2023 with a time interval of 24 hours extracted from the ERA5 reanalysis dataset. All models are trained on 1979-2017 data and vaidated on 2017-2022 data. The evaluations presented here are done on the 2023 data.




\textbf{Baselines.} We extensively compare our proposed PhyDA with several state-of-the-art (SOTA) data assimilation approaches. FNP~\citep{chen2024fnp}, leveraging efficient module design and the flexible structure of neural processes, achieves SOTA performance for arbitrary-resolution data assimilation. VAE-VAR~\citep{xiao2025vae} employs a variational autoencoder \citep{kingma2013auto} to model background error distributions. SDA~\citep{rozet2023score} introduces score-based data assimilation for trajectory inference, and SLAM~\citep{qu2024deep} builds upon SDA by incorporating multimodal observational data.

\textbf{Model training and evaluation.} 
We evaluate the performance of models by calculating the overall mean square error (MSE), mean absolute error (MAE), 
and the latitude-weighted root mean square error (RMSE) which is a statistical metric widely used in geospatial analysis and atmospheric science, following the setting of FNP \citep{chen2024fnp}. 
Given the estimate $\hat{x}_{h,w,c}$ and its ground truth $x_{h,w,c}$ for the $c$-th channel, the RMSE is defined as:
{\small
\begin{equation}
\operatorname{RMSE}(c) = \sqrt{\frac{1}{H\cdot W}\sum\nolimits_{h,w} H \frac{\operatorname{cos}(\alpha_{h,w})}{\sum_{h'=1}^{H} \operatorname{cos}(\alpha_{h',w})}(x_{h,w,c} - \hat{x}_{h,w,c})^{2}}, 
\end{equation}
}

where $H$ and $W$ represent the number of grid points in the longitudinal and latitudinal directions, respectively, and $\alpha_{h,w}$ is the latitude of point $(h,w)$.

\textbf{Physical evaluation.} As mentioned before, purely data-driven models may generate physically implausible outputs. To evaluate the physical coherence of the generated state, we leverage the physics-based metric, Spectral Divergence (SpecDiv), from ChaosBench \citep{nathaniel2024chaosbench} that measures the deviation between the power spectra of assimilated state and ground truth.  
SpecDiv follows principles from Kullback–Leibler (KL) divergence where we compute the expectation of the log ratio between target $S^\prime(\omega)$ and prediction $\hat{S}^\prime(\omega)$ spectra, and is defined as:
\begin{equation}
    \mathcal{M}_{SpecDiv} = \sum_{\omega} S^{\prime}(\omega) \cdot \log(S^{\prime}(\omega) / \hat{S}^{\prime}(\omega)),
    \label{eq:specdiv}
\end{equation}
where $\omega \in \mathbf{\Omega}$, and $\mathbf{\Omega}$ is the set of all scalar wavenumbers from 2D Fourier transform. More details can be seen in Appendix.

\textbf{Treatment of sparse observations.}
Acknowledging the multidimensional nature of the state vector and that most meteorological observations are co-located horizontally (longitude and latitude), we opt for a simplified setting in the conditioning of sparse observations.
In this scenario, the observational data is $h$ sampled columns of
the ground truth state vector with $6 \times 13 +4$ values in each column: $y \in \mathbb{R}^{(6 \times 13 +4)\times h}$.
The mask is simplified to a 2-D mask $h_s \in \mathbb{R}^{121\times 240}$ which is broadcast to other value dimensions when doing point-wise multiplication with the state vector.
The normalization is also applied independently over 2D horizonal slices for each variable and pressure level.

\begin{table}[t]
  \caption{Quantitative performance comparison for deep learning methods for data assimilation. The best performance are shown in \textbf{bold} while the second best is \underline{underscored}. The baseline results are average of 5 parallel experiments. We show the MSE and MAE over all variables and RMSE of part of the variables. Red color indicates \better{\textbf{physical metric SpecDiv}} of z500 and indicates the improved assimilation results compared to that with background.}
  \label{tab:main_experiments}
  \centering
  \resizebox{\linewidth}{!}{
  \setlength{\extrarowheight}{1pt}
  \begin{tabular}{l|c|cc|cccccccc}
    \toprule
    \multirow{2}{*}{Model} & \multirow{2}{*}{SpecDiv $\downarrow$} & \multirow{2}{*}{MSE($10^{-2}$)$\downarrow$} & \multirow{2}{*}{MAE$(10^{-2})$$\downarrow$} & \multicolumn{8}{c}{RMSE$\downarrow$} \\
    & & & & z500 & t850 & t2m & u10 & v10 & u500 & v500 & q700 ($10^{-4}$) \\
    \midrule
    Background & 0.153 & 2.88 & 8.61 & 45.455 & 0.7200 & 0.7790 & 0.9336 & 0.9645 & 1.7278 & 1.7535 & 6.7220 \\
    \midrule
    \midrule

    Adas \citep{chen2023adas} &\better{0.116 \small{$\downarrow$ 24.2\%}} & 2.31 & 7.65 & 30.100 & \underline{0.6750} & 0.7350 & 0.8400 & 0.8600 & 1.4950 & 1.4900 & 6.5400 \\
    ConvCNP \citep{gordon2019convcnp} &\better{0.125 \small{$\downarrow$ 18.3\%}} & 2.49 & 7.98 & 31.253 & 0.6944 & 0.7662 & 0.8334 & 0.8553 & 1.5770 & 1.5876 & 6.5717 \\ 
    FNP \citep{chen2024fnp}&\better{0.063  \small{$\downarrow$ 58.8\%}}& \underline{2.30}  & \underline{7.54} & 28.500 & 0.6985 & 0.7100 & 0.7650 & \underline{0.7650} & \underline{1.4350} & 1.4600 & \underline{6.4698} \\

    VAE-VAR \citep{xiao2025vae} &\better{\underline{0.052} \small{$\downarrow$ 66.0\%}} & 2.31 & 7.60 & \textbf{27.000} & 0.6970 & \underline{0.7050} & \underline{0.7560} & 0.7770 & 1.4500 & \underline{1.4500} & 6.4700 \\
    SDA  \citep{rozet2023score}   &\better{0.117 \small{$\downarrow$ 23.5\%}}  & 2.65 & 8.02 & 38.000 & 0.7100 & 0.7500 & 0.8800 & 0.9100 & 1.6500 & 1.7000 & 6.6100 \\
    SLAM  \citep{qu2024deep}  &\better{0.091 \small{$\downarrow$ 40.5\%}} & 2.55 & 7.94 & 32.500 & 0.7020 & 0.7300 & 0.8000 & 0.7800 & 1.5000 & 1.4700 & 6.5000 \\

    \midrule
  
    PhyDA &\textbf{\better{0.031 \small{$\downarrow$ 79.7\%}}} &\textbf{2.28}  &\textbf{7.53 }  &\underline{26.866}   &\textbf{0.6684}   &\textbf{0.7033}  &\textbf{0.7549} &\textbf{0.7638}&\textbf{1.4258}&\textbf{1.4462}&\underline{6.4692}\\
    Floating Error & $\pm 0.008$&$\pm$0.03 &$\pm$0.04 &$\pm$ 0.005 &$\pm$ 0.0003 &$\pm$ 0.0006 &$\pm$0.0010 &$\pm$ 0.0006 &$\pm$0.0098 & $\pm$0.0098&$\pm$0.0004\\
    \bottomrule
  \end{tabular}
  }
\end{table}

\subsection{Experiment Results}
We validate the performance of models by assimilating 2\% observations with resolutions of 1.5°, respectively, onto a 24-hour forecast background with 1.5° resolution. 
Table~\ref{tab:main_experiments} provides a quantitative comparison of the analysis errors between PhyDA and other models. 
The first row corresponds to the error level of the background. 
PhyDA achieves SOTA results (indicated in bold) in terms of overall SpecDiv, MSE, MAE, and RMSE metrics for the majority of variables. 

Specifically, it obtains the lowest SpecDiv, indicating the highest degree of physical consistency in the reconstructed z500 field. This is a significant improvement over the background field and outperforms all baselines by a large margin.
In terms of numerical accuracy, PhyDA achieves the lowest MSE and lowest MAE across all variables. For variable-specific RMSE, PhyDA outperforms other models on most key variables. These results demonstrate PhyDA’s strong capability to reconstruct accurate and physically consistent atmospheric states, particularly under sparse observation scenarios.

\begin{figure}[htbp]
    \centering
    \includegraphics[width=0.95\linewidth]{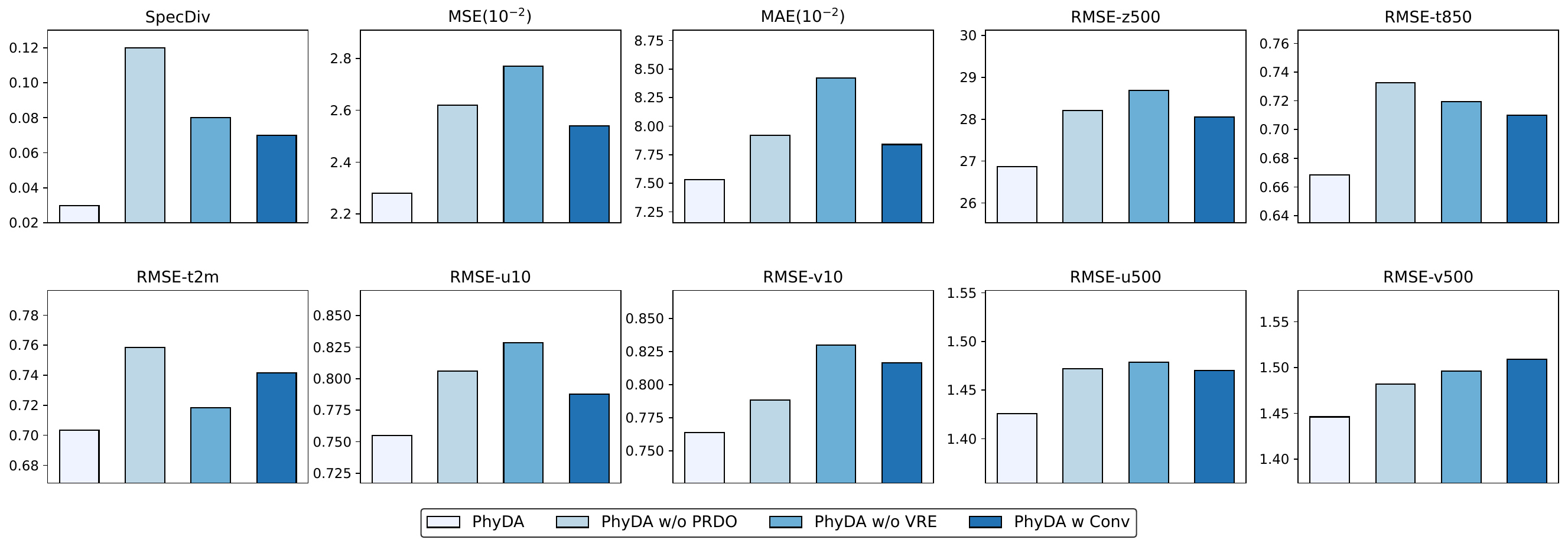}
    \caption{Ablation study. 
    }
     \vspace{-10pt}
    \label{fig:ablation}
\end{figure}
\textbf{Ablation Study.}
To verify the effectiveness of the proposed method, as shown in Figure~\ref{fig:ablation}, we conducted detailed ablation experiments.
We introduce the following model variants: 
(1) \textbf{PhyDA w/o PRDO}, we remove the Physically Regularized Diffusion Objective (PRDO) and leverage the original training objective of score-based diffusion model.
(2) \textbf{PhyDA w/o VRE}, we remove the Virtual Reconstruction Encoder (VRE) thus remove the conditional information added in the inference process.
(3) \textbf{PhyDA w Conv}, we replace VRE with a common-used convlutional encoder.

The results of the ablation experiments show that remove either PRDO or VRE will reduce the assimilation performance, further proving the rationality of the components proposed by PhyDA. 
When we remove PRDO, the SpecDiv dropped significantly, demonstrating the effectiveness of PRDO in enforcing physical coherence among multiple atmospheric variables.
Furthermore, experiment results in (2) and (3) demonstrate that additional condition information can boost the generative quality of PhyDA, where the well-designed VRE can better capture intricate physical dependencies from sparse observations, which achieves satisfactory results in real-world data assimilation tasks.

\begin{figure}[htbp]
    \centering
     \vspace{-5pt}
    \includegraphics[width=0.95\linewidth]{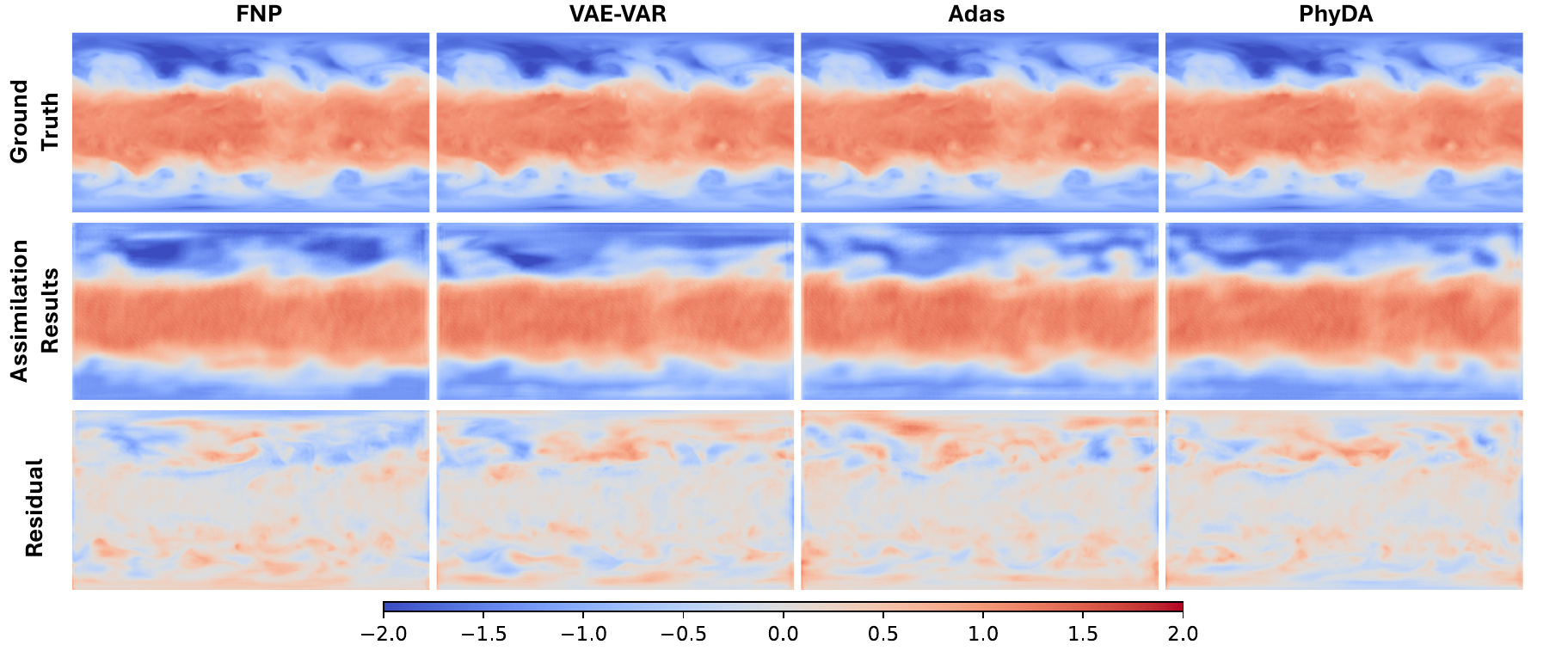}
    \vspace{-8pt}
    \caption{Realistic case where we use PhyDA to assimilate t@250hPa.}
     \vspace{-5pt}
    \label{fig:visualization}
\end{figure}
\textbf{Visualization.}
The visualization of Figure~\ref{fig:visualization} reveals that when applied to real-world data assimilation tasks, there is blurring occurring at different spatial scales, resulting in the loss of extreme values, as indicated in the red box.
We compare the assimilation results from sota models including FNP, VAE-VAR, Adas. The residuals shown in the third row indicate that PhyDA achieves the best performance.
Our model, which incorporates physical constraints, provides clearer assimilation retaining rules in atmospheric dynamics without the need for further adjusting.

\begin{figure}[htbp]
    \centering
    \vspace{-5pt}
    \includegraphics[width=\linewidth]{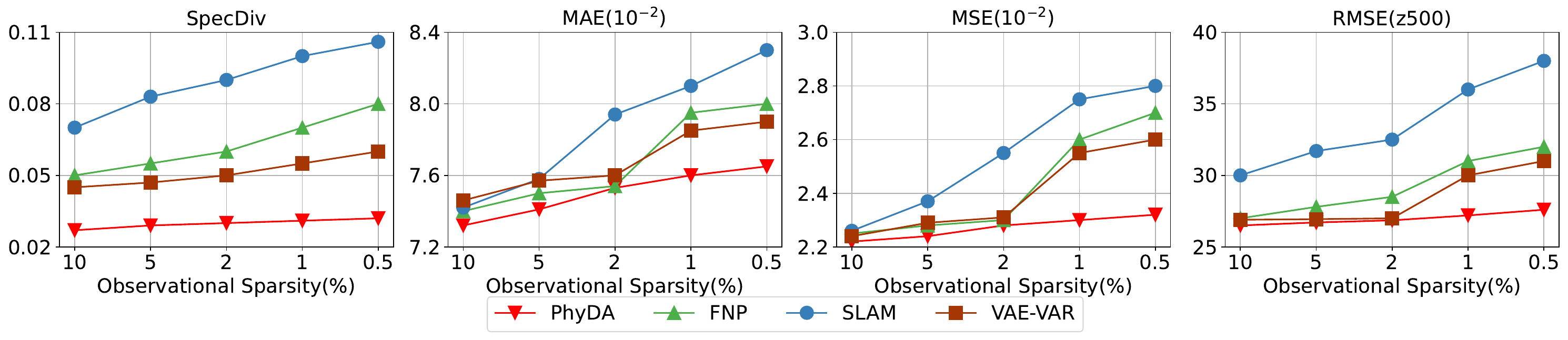}
    \vspace{-10pt}
    \caption{Model performance with multiple observational sparsity ($\{10\%, 5\%, 2\%, 1\%, 0.5\%\}$).}
    \vspace{-5pt}
    \label{fig:multi_sparsity}
\end{figure}
\textbf{Increasing observational sparsity.}
 In this part, we analyze the model performance across varying observational sparsity, including $ \{10\%, 5\%, 2\%, 1\%, 0.5\%\}$. Figure~\ref{fig:multi_sparsity} showcases the SpecDiv, MAE, MSE and RMSE for z500.    
 The PhyDA model demonstrates the most stable performance among MAE, MSE and RMSE metrics, particularly in extreme observational sparsity ($0.5\%$). In contrast, previous state-of-the-art models exhibit notable performance degradation as the observational sparsity increases. The SpecDiv performance of PhyDA, on the other hand, remains relatively stable and robust, demonstrating its efficacy in leveraging PRDO to enforce physical coherence, which can also be attributed to its ability to capture intricate correlations effectively even with few observations, making it highly suited for real-world data assimilation tasks.

\section{Related Work}

\subsection{Physics-Guided Machine Learning of Spatiotemporal Dynamics}

Physics-guided machine learning aims to incorporate known scientific laws—often expressed as partial differential equations (PDEs)—into data-driven models to improve generalization, stability, and interpretability \citep{karniadakis2021physics}. The most common strategy is to add physics-based regularization to the loss function, as in Physics-Informed Neural Networks (PINNs) \citep{raissi2019physics}, which directly penalize the residuals of governing PDEs. Extensions of this idea have been applied to fluid dynamics \citep{wang2020towards}, environmental forecasting \citep{daw2022physics}, and molecular systems \citep{zhang2018deep}. However, PINN-based approaches are known to suffer from training instability and high computational cost, particularly when enforcing high-order operators (e.g., Navier–Stokes equations) or learning in high-dimensional, chaotic systems.
More recent efforts embed physical structure at the architectural level \citep{long2018hybridnet, kashinath2020enforcing}, or use Graph Neural Networks and Fourier layers to better respect conservation laws \citep{li2020fourier, li2020neural}.  
Although \citep{bastek2024physics} offers a promising solution
by integrating physic laws with DDPM \citep{ho2020denoising}, such training instability and high computational cost still hinder its direct application in DA tasks.

Our work bridges these gaps by introducing a physics-regularized score objective that jointly minimizes PDE residuals and standard score matching in a theoretically grounded way. Notably, embedding governing equations into score-based diffusion models is non-trivial, as each component—from noise scheduling to reverse-time sampling—is derived from strict stochastic theory. We show that PhyDA improves both physical fidelity and generalization in data assimilation tasks.

\subsection{Deep Learning Approaches for Data Assimilation}

Data assimilation (DA) plays a critical role in Earth system modeling by combining prior numerical forecasts with real-time observational data to estimate the system state. Traditional DA methods (e.g., 3D-Var, 4D-Var, EnKF~\citep{le1986variational, courtier1998ecmwf, tr2006accounting}) rely heavily on linear approximations, Gaussian assumptions, and adjoint solvers, limiting their ability to capture nonlinear dynamics or scale to large systems.
Recent advances in deep learning (DL) have introduced deep learning-based alternatives that aim to improve scalability and accuracy. Several studies demonstrate that DL models trained on reanalysis data can emulate or surpass traditional numerical weather prediction systems \citep{lam2023learning, bi2022pangu}. Explorations have also been made to train DL models for data assimilation tasks. FNP \citep{chen2024fnp} assimilates observations with varying resolutions leveraging carefully designed modules of neural process. VAE-Var \citep{xiao2025vae} utilizes a variational autoencoder \citep{kingma2013auto} to learn the background error distribution. Specifically, diffusion models have gained traction for reconstructing dynamic states from partial observations. Score-based Data Assimilation (SDA) \citep{rozet2023score} leverages reverse-time score-based models to estimate high-dimensional fields, extending to geophysical systems like two-layer quasi-geostrophic flows \citep{rozet2023scoretwo}. DiffDA \citep{huang2024diffda} further explores autoregressive diffusion-based forecasting with pseudo-observation assimilation.

However, these approaches typically treat the generative process as purely data-driven and do not incorporate domain-specific physical constraints in atmospheirc systems. Moreover, they often assume access to dense observations during training, limiting their robustness under sparse or irregular observational coverage. In contrast, our method explicitly targets these limitations by integrating physical constraints into the generative score-matching objective, and learning to encode sparse observations into spatially continuous, physically structured latent space.

\section{Conclusion and Future Works}

In this paper, we highlight the significance of preserving physical coherence across coupled atmospheric variables in real-world data assimilation tasks.
To address this problem, we introduce PhyDA, an innovative physics-guided data assimilation method, which effectively integrates physical constraints into score-based diffusion models, thereby ensuring the physical coherence of assimilated atmospheric variables. 
At the core of PhyDA are two key innovative components: the Physically Regularized Diffusion Objective (PRDO) and the Virtual Reconstruction Encoder (VRE).  PRDO guides the model to generate analysis fields that conform to physical laws by incorporating physical constraints into the generative process in a theoretically grounded way by penalizing deviations from governing partial differential equations (PDEs), alongside the standard score-matching loss.
Meanwhile, VRE is capable of extracting structured latent representations from sparse observational data, further enhancing the model's ability to infer complete and physically consistent states.
As a result,  PhyDA bridges data-driven generation with domain-specific physical principles, ensuring both accuracy and plausibility in assimilation outputs. 
Experimental results fully validate the effectiveness of the PhyDA method, which significantly improves the physical plausibility of these results.

For future works, we aim to extend our framework to integrate multimodal observational data, such as satellite imagery and radar measurements, to further improve the accuracy and robustness of data assimilation. Additionally,  we will investigate end-to-end weather forecasting models that directly combine assimilation and prediction within a unified learning framework, potentially advancing the state of real-time atmospheric modeling.

\bibliographystyle{plain}
\bibliography{neurips_2025}

\begin{thebibliography}{10}

\bibitem{bannister2017review}
Ross~N Bannister.
\newblock A review of operational methods of variational and ensemble-variational data assimilation.
\newblock {\em Quarterly Journal of the Royal Meteorological Society}, 143(703):607--633, 2017.

\bibitem{bastek2024physics}
Jan-Hendrik Bastek, WaiChing Sun, and Dennis~M Kochmann.
\newblock Physics-informed diffusion models.
\newblock {\em arXiv preprint arXiv:2403.14404}, 2024.

\bibitem{bi2022pangu}
Kaifeng Bi, Lingxi Xie, Hengheng Zhang, Xin Chen, Xiaotao Gu, and Qi~Tian.
\newblock Pangu-weather: A 3d high-resolution model for fast and accurate global weather forecast.
\newblock {\em arXiv preprint arXiv:2211.02556}, 2022.

\bibitem{brandstetter2022clifford}
Johannes Brandstetter, Daniel~E Worrall, and Max Welling.
\newblock Clifford neural networks for learning stable dynamical systems.
\newblock In {\em International Conference on Learning Representations (ICLR)}, 2022.

\bibitem{carrassi2018data}
Alberto Carrassi, Marc Bocquet, Laurent Bertino, and Geir Evensen.
\newblock Data assimilation in the geosciences: An overview of methods, issues, and perspectives.
\newblock {\em Wiley Interdisciplinary Reviews: Climate Change}, 9(5):e535, 2018.

\bibitem{chen2023fengwu}
Kang Chen, Tao Han, Junchao Gong, Lei Bai, Fenghua Ling, Jing-Jia Luo, Xi~Chen, Leiming Ma, Tianning Zhang, Rui Su, et~al.
\newblock Fengwu: Pushing the skillful global medium-range weather forecast beyond 10 days lead.
\newblock {\em arXiv preprint arXiv:2304.02948}, 2023.

\bibitem{chen2023adas}
Kun Chen, Lei Bai, Fenghua Ling, Peng Ye, Tao Chen, Jing-Jia Luo, Hao Chen, Yi~Xiao, Kang Chen, Tao Han, et~al.
\newblock Towards an end-to-end artificial intelligence driven global weather forecasting system.
\newblock {\em arXiv preprint arXiv:2312.12462}, 2023.

\bibitem{chen2024fnp}
Kun Chen, Peng Ye, Hao Chen, Tao Han, Wanli Ouyang, Tao Chen, LEI BAI, et~al.
\newblock Fnp: Fourier neural processes for arbitrary-resolution data assimilation.
\newblock {\em Advances in Neural Information Processing Systems}, 37:137847--137872, 2024.

\bibitem{courtier1998ecmwf}
Philippe Courtier, E~Andersson, W~Heckley, D~Vasiljevic, M~Hamrud, A~Hollingsworth, F~Rabier, M~Fisher, and J~Pailleux.
\newblock The ecmwf implementation of three-dimensional variational assimilation (3d-var). i: Formulation.
\newblock {\em Quarterly Journal of the Royal Meteorological Society}, 124(550):1783--1807, 1998.

\bibitem{daw2022physics}
Arka Daw, Anuj Karpatne, William~D Watkins, Jordan~S Read, and Vipin Kumar.
\newblock Physics-guided neural networks (pgnn): An application in lake temperature modeling.
\newblock In {\em Knowledge Guided Machine Learning}, pages 353--372. Chapman and Hall/CRC, 2022.

\bibitem{dhariwal2021diffusion}
Prafulla Dhariwal and Alexander Nichol.
\newblock Diffusion models beat gans on image synthesis.
\newblock {\em Advances in neural information processing systems}, 34:8780--8794, 2021.

\bibitem{evensen1994sequential}
Geir Evensen.
\newblock Sequential data assimilation with a nonlinear quasi-geostrophic model using monte carlo methods to forecast error statistics.
\newblock {\em Journal of Geophysical Research: Oceans}, 99(C5):10143--10162, 1994.

\bibitem{gordon2019convcnp}
Jonathan Gordon, Wessel~P Bruinsma, Andrew~YK Foong, James Requeima, Yann Dubois, and Richard~E Turner.
\newblock Convolutional conditional neural processes.
\newblock {\em arXiv preprint arXiv:1910.13556}, 2019.

\bibitem{gupta2024photorealistic}
Agrim Gupta, Lijun Yu, Kihyuk Sohn, Xiuye Gu, Meera Hahn, Fei-Fei Li, Irfan Essa, Lu~Jiang, and Jos{\'e} Lezama.
\newblock Photorealistic video generation with diffusion models.
\newblock In {\em European Conference on Computer Vision}, pages 393--411. Springer, 2024.

\bibitem{hersbach2019era5}
Hans Hersbach, Bill Bell, Paul Berrisford, G~Biavati, Andr{\'a}s Hor{\'a}nyi, Joaqu{\'\i}n Mu{\~n}oz~Sabater, Julien Nicolas, C~Peubey, R~Radu, I~Rozum, et~al.
\newblock Era5 monthly averaged data on single levels from 1979 to present.
\newblock {\em Copernicus Climate Change Service (C3S) Climate Data Store (CDS)}, 10:252--266, 2019.

\bibitem{ho2020denoising}
Jonathan Ho, Ajay Jain, and Pieter Abbeel.
\newblock Denoising diffusion probabilistic models.
\newblock {\em Advances in neural information processing systems}, 33:6840--6851, 2020.

\bibitem{holton2012dynamic}
James~R. Holton and Gregory~J. Hakim.
\newblock {\em An Introduction to Dynamic Meteorology}.
\newblock Academic Press, 2012.

\bibitem{huang2024diffda}
Langwen Huang, Lukas Gianinazzi, Yuejiang Yu, Peter~D Dueben, and Torsten Hoefler.
\newblock Diffda: a diffusion model for weather-scale data assimilation.
\newblock {\em arXiv preprint arXiv:2401.05932}, 2024.

\bibitem{karniadakis2021physics}
George~Em Karniadakis, Ioannis~G Kevrekidis, Lu~Lu, Paris Perdikaris, Sifan Wang, and Liu Yang.
\newblock Physics-informed machine learning.
\newblock {\em Nature Reviews Physics}, 3(6):422--440, 2021.

\bibitem{kashinath2020enforcing}
Karthik Kashinath, Philip Marcus, et~al.
\newblock Enforcing physical constraints in cnns through differentiable pde layer.
\newblock In {\em ICLR 2020 Workshop on Integration of Deep Neural Models and Differential Equations}, 2020.

\bibitem{kingma2013auto}
Diederik~P Kingma, Max Welling, et~al.
\newblock Auto-encoding variational bayes, 2013.

\bibitem{lam2023learning}
Remi Lam, Alvaro Sanchez-Gonzalez, Matthew Willson, Peter Wirnsberger, Meire Fortunato, Ferran Alet, Suman Ravuri, Timo Ewalds, Zach Eaton-Rosen, Weihua Hu, et~al.
\newblock Learning skillful medium-range global weather forecasting.
\newblock {\em Science}, 382(6677):1416--1421, 2023.

\bibitem{le1986variational}
Fran{\c{c}}ois-Xavier Le~Dimet and Olivier Talagrand.
\newblock Variational algorithms for analysis and assimilation of meteorological observations: theoretical aspects.
\newblock {\em Tellus A: Dynamic Meteorology and Oceanography}, 38(2):97--110, 1986.

\bibitem{li2020fourier}
Zongyi Li, Nikola Kovachki, Kamyar Azizzadenesheli, Burigede Liu, Kaushik Bhattacharya, Andrew Stuart, and Anima Anandkumar.
\newblock Fourier neural operator for parametric partial differential equations.
\newblock {\em arXiv preprint arXiv:2010.08895}, 2020.

\bibitem{li2020neural}
Zongyi Li, Nikola Kovachki, Kamyar Azizzadenesheli, Burigede Liu, Kaushik Bhattacharya, Andrew Stuart, and Anima Anandkumar.
\newblock Neural operator: Graph kernel network for partial differential equations.
\newblock {\em arXiv preprint arXiv:2003.03485}, 2020.

\bibitem{long2018hybridnet}
Yun Long, Xueyuan She, and Saibal Mukhopadhyay.
\newblock Hybridnet: integrating model-based and data-driven learning to predict evolution of dynamical systems.
\newblock In {\em Conference on robot learning}, pages 551--560. PMLR, 2018.

\bibitem{lorenc1986analysis}
Andrew~C Lorenc.
\newblock Analysis methods for numerical weather prediction.
\newblock {\em Quarterly Journal of the Royal Meteorological Society}, 112(474):1177--1194, 1986.

\bibitem{nathaniel2024chaosbench}
Juan Nathaniel, Yongquan Qu, Tung Nguyen, Sungduk Yu, Julius Busecke, Aditya Grover, and Pierre Gentine.
\newblock Chaosbench: A multi-channel, physics-based benchmark for subseasonal-to-seasonal climate prediction.
\newblock {\em arXiv preprint arXiv:2402.00712}, 2024.

\bibitem{qu2024deep}
Yongquan Qu, Juan Nathaniel, Shuolin Li, and Pierre Gentine.
\newblock Deep generative data assimilation in multimodal setting.
\newblock In {\em Proceedings of the IEEE/CVF Conference on Computer Vision and Pattern Recognition}, pages 449--459, 2024.

\bibitem{raissi2019physics}
Maziar Raissi, Paris Perdikaris, and George~E Karniadakis.
\newblock Physics-informed neural networks: A deep learning framework for solving forward and inverse problems involving nonlinear partial differential equations.
\newblock {\em Journal of Computational physics}, 378:686--707, 2019.

\bibitem{rasp2020weatherbench}
Stephan Rasp, Peter~D Dueben, Sebastian Scher, Jonathan~A Weyn, Soukayna Mouatadid, and Nils Thuerey.
\newblock Weatherbench: A benchmark dataset for data-driven weather forecasting.
\newblock {\em Quarterly Journal of the Royal Meteorological Society}, 146(730):1745--1769, 2020.

\bibitem{rozet2023score}
Fran{\c{c}}ois Rozet and Gilles Louppe.
\newblock Score-based data assimilation.
\newblock {\em Advances in Neural Information Processing Systems}, 36:40521--40541, 2023.

\bibitem{rozet2023scoretwo}
Fran{\c{c}}ois Rozet and Gilles Louppe.
\newblock Score-based data assimilation for a two-layer quasi-geostrophic model.
\newblock {\em arXiv preprint arXiv:2310.01853}, 2023.

\bibitem{ruhling2024probablistic}
Salva R{\"u}hling~Cachay, Brian Henn, Oliver Watt-Meyer, Christopher~S Bretherton, and Rose Yu.
\newblock Probablistic emulation of a global climate model with spherical dyffusion.
\newblock {\em Advances in Neural Information Processing Systems}, 37:127610--127644, 2024.

\bibitem{tr2006accounting}
Yannick Tr'emolet.
\newblock Accounting for an imperfect model in 4d-var.
\newblock {\em Quarterly Journal of the Royal Meteorological Society: A journal of the atmospheric sciences, applied meteorology and physical oceanography}, 132(621):2483--2504, 2006.

\bibitem{wallace2006atmospheric}
John~M. Wallace and Peter~V. Hobbs.
\newblock {\em Atmospheric Science: An Introductory Survey}.
\newblock Academic Press, 2006.

\bibitem{wang2020towards}
Rui Wang, Karthik Kashinath, Mustafa Mustafa, Adrian Albert, and Rose Yu.
\newblock Towards physics-informed deep learning for turbulent flow prediction.
\newblock In {\em Proceedings of the 26th ACM SIGKDD international conference on knowledge discovery \& data mining}, pages 1457--1466, 2020.

\bibitem{wang2021physics}
Rui Wang and Rose Yu.
\newblock Physics-guided deep learning for dynamical systems: A survey.
\newblock {\em arXiv preprint arXiv:2107.01272}, 2021.

\bibitem{wu2024transolver}
Haixu Wu, Huakun Luo, Haowen Wang, Jianmin Wang, and Mingsheng Long.
\newblock Transolver: A fast transformer solver for pdes on general geometries.
\newblock {\em arXiv preprint arXiv:2402.02366}, 2024.

\bibitem{xiao2025vae}
Yi~Xiao, Qilong Jia, Kun Chen, Lei Bai, and Wei Xue.
\newblock Vae-var: Variational autoencoder-enhanced variational methods for data assimilation in meteorology.
\newblock In {\em The Thirteenth International Conference on Learning Representations}, 2025.

\bibitem{xu2024generalizing}
Wanghan Xu, Fenghua Ling, Wenlong Zhang, Tao Han, Hao Chen, Wanli Ouyang, and Lei Bai.
\newblock Generalizing weather forecast to fine-grained temporal scales via physics-ai hybrid modeling.
\newblock {\em arXiv preprint arXiv:2405.13796}, 2024.

\bibitem{zhang2018deep}
Linfeng Zhang, Jiequn Han, Han Wang, Roberto Car, and Weinan E.
\newblock Deep potential molecular dynamics: a scalable model with the accuracy of quantum mechanics.
\newblock {\em Physical review letters}, 120(14):143001, 2018.

\end{thebibliography}


\end{document}